%% file: main.tex
\documentclass[10pt,twocolumn,letterpaper]{article}

\usepackage[final]{cvpr}      % To produce the REVIEW version review, final, preprint
\usepackage{multirow}
\usepackage{pifont}
\usepackage{makecell}
\input{preamble}
\definecolor{cvprblue}{rgb}{0.21,0.49,0.74}
\usepackage[pagebackref,breaklinks,colorlinks,allcolors=cvprblue]{hyperref}

\def\paperID{*****} % *** Enter the Paper ID here
\def\confName{CVPR}
\def\confYear{2026}

\title{AuroraEdge-V-2B: A Faster And Stronger Edge Visual Large Language Model}

\author{
  Xiang Chen \\
  Independent Researcher, Hangzhou, China \\
  {\tt\small chenxiang101@zuaa.zju.edu.cn}
}

\begin{document}
\maketitle
\input{sec/0_abstract}    
\input{sec/1_intro}
\input{sec/2_related}
\input{sec/3_model}
\input{sec/4_imple}    
\input{sec/5_result}
\input{sec/6_ablation}
\input{sec/7_limit}
\input{sec/8_conclusion}
\input{sec/X_suppl}
{
    \small
    \bibliographystyle{ieeenat_fullname}
    \bibliography{main}
}


\end{document}

%% file: sec/0_abstract.tex
\begin{abstract}
Recently, due to the advancement of multimodal technology, people are attempting to use visual 
large language models (VLLMs) in industrial production. Many deep learning models (DLMs) deployed 
in the production environment are gradually being replaced by VLLMs. Compared with DLMs, VLLMs have 
some advantages in industrial applications: (1) Their strong generalization ability enables them to perform well across a wide range of tasks. 
(2) They are flexible and can deal with unfamiliar samples through context learning quickly. 
However, VLLMs also have obvious drawbacks: (1) VLLMs do not perform as well as custom-developed DLMs
in specific domains. (2) The number of parameters in VLLMs is generally quite large, and their 
deployment requires substantial computational resources. (3) VLLMs generally operate much slower than DLMs, 
making real-time response challenging to achieve. To better utilize VLLMs in industrial applications, we 
introduce AuroraEdge-V-2B in this work, a compact, robust, and high-speed VLLM designed for edge deployment. 
To make the model run faster, we also propose a compression-fusion method to improve inference efficiency. 
AuroraEdge-V-2B has the following notable features: (1) \textbf{Easy deployment and faster:} It has only 
2B parameters and is highly suitable for edge deployment, offering better real-time performance. 
(2) \textbf{Fewer visual tokens and cheaper:} It significantly reduces the number of visual tokens in the 
decoding process, thereby reducing the floating-point operations by half during inference and making it cheaper 
to use. (3) \textbf{Strong performance:} It gets a higher score on 9 benchmarks than models with the same number of parameter
(e.g., Qwen2-VL-2B, Qwen2.5-VL-3B, InternVL-2.5-2B). 
\end{abstract}

%% file: sec/1_intro.tex
\section{Introduction}
Over the past decade, the industrial sector has witnessed widespread adoption of DLMs within production processes. 
Nevertheless, despite their efficacy in reducing labor costs, the development of these models often incurs high costs and long cycles.
Moreover, such models are characteristically confined to operating on data within the domain of the training samples, exhibiting a circumscribed range of capabilities.

\begin{figure}
    \centering
    \includegraphics[scale=0.24]{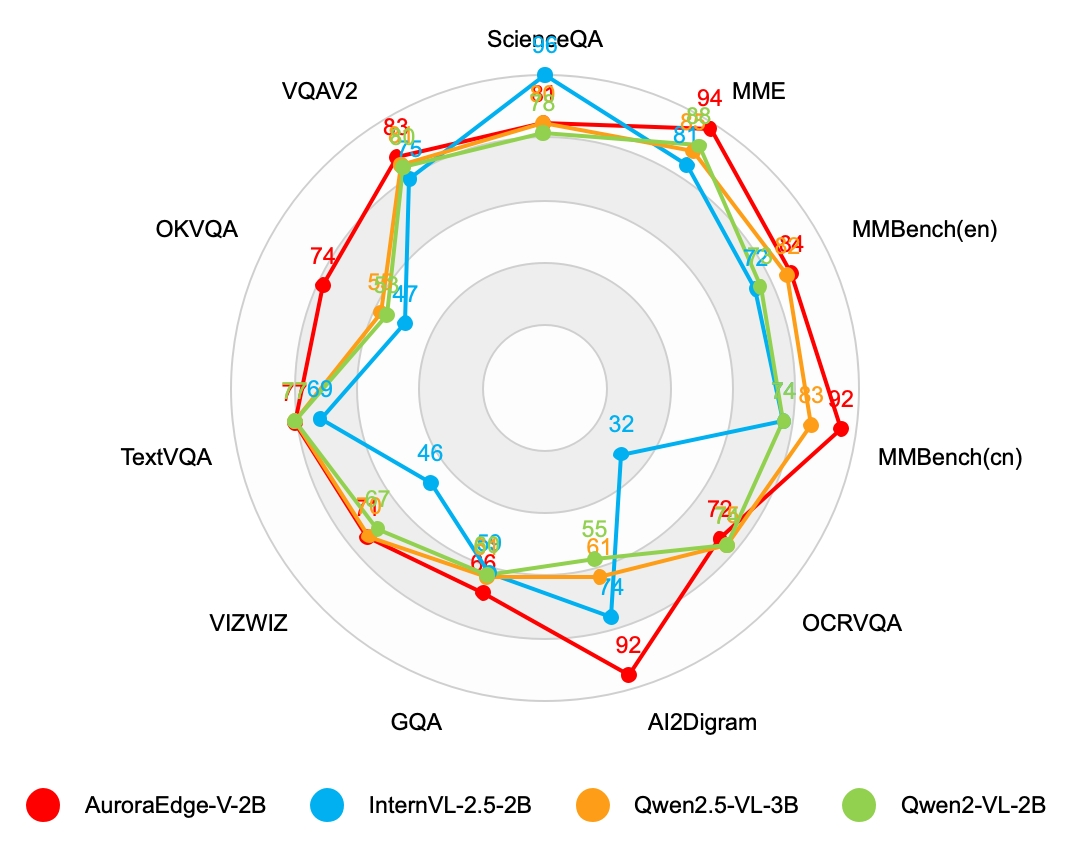}
    \caption{AuroraEdge-V-2B achieves high scores on 11 benchmarks, outperforming other state-of-the-art (SOTA) models with the same parameter scale on 9 of them.}
    \label{fig:res_01}
\end{figure}

Driven by the enhanced capabilities of VLLMs(\cite{ref1,ref3,ref4,ref5,ref47,ref48,ref49,ref50,ref51,ref52}), a paradigm shift is underway, with engineers increasingly favoring VLLMs over traditional DLMs for industrial deployment.
Compared with DLMs, VLLMs offer the following advantages: 
(1) Their strong generalization ability enables them to perform well across a wide range of tasks. (2) They are flexible and can deal with new samples through context learning quickly. 
The most popular method for building VLLMs is the LLaVA architecture\cite{ref1,ref3,ref4,ref5}, which mainly consists of three key components: 
the visual encoder (VE), the projector, and the LLM backbone.

\begin{figure}
    \centering
    \includegraphics[scale=0.22]{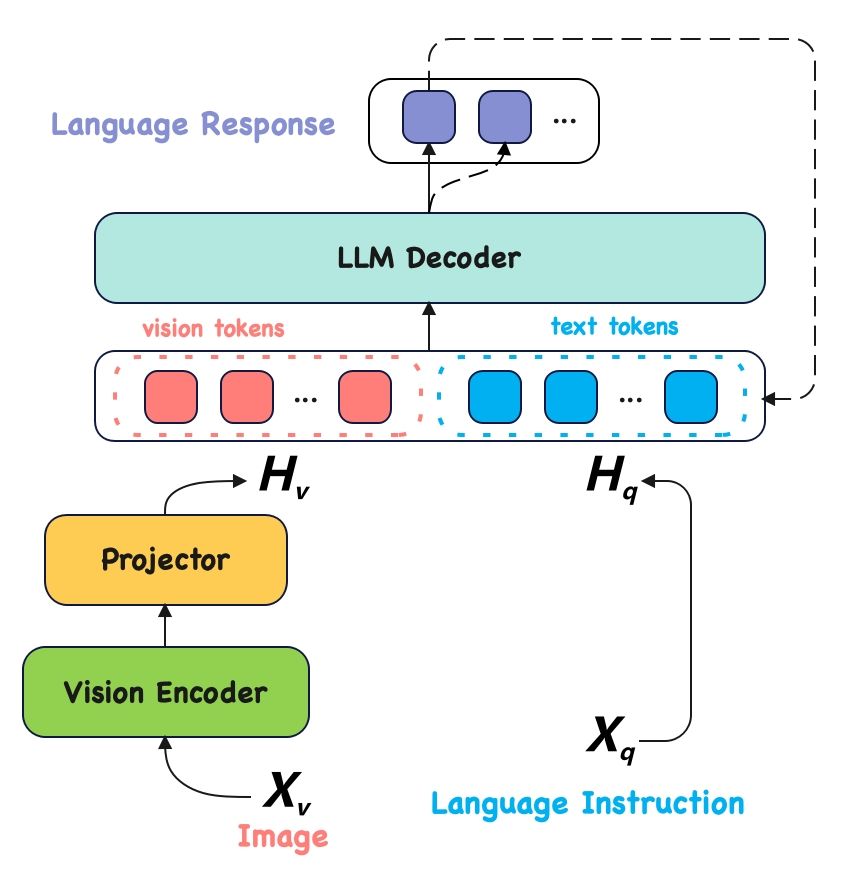}
    \caption{The LLaVA architecture.}
    \label{fig:llava_01}
\end{figure}

The LLaVA architecture(see Figure \ref{fig:llava_01})\cite{ref1,ref3,ref4,ref5} provides a highly efficient framework for constructing VLLMs, particularly when the LLM backbone and the VE are initialized with pre-trained weights.
Due to the high real-time requirements of industrial applications, we generally opt for LLMs with fewer parameters to construct VLLMs, thereby facilitating the edge deployment.
There is a wide array of small pretrained LLMs available, such as Qwen2.5 series (1.5B, 3B, 7B)\cite{ref27}, the Gemma series (1B, 2B, 4B)\cite{ref28},  among others.
Similarly, numerous outstanding works have contributed powerful pre-trained VEs,including CLIP\cite{ref8,ref9} 
and SigLIP\cite{ref10}, etc. Additionally, dynamic resolution processors\cite{ref11,ref5} are also widely used to enhance the performance of VLLMs
on tasks like document understanding and Optical Character Recognition (OCR). 

However, VLLMs still possess significant shortcomings. They lag behind DLMs in terms of performance and latency within specific domains,
and their larger parameter counts necessitates substantial computational resources for deployment.
Unlike LLMs, VLLMs produce a large number of visual tokens for the decoder, which significantly increases their computational burden. For example, the VE 
employed in LLaVa-1.5\cite{ref4}, which is CLIP ViT-L/336px\cite{ref8}, encode each image into 576 visual tokens, whereas a simple 
text query is typically represented by only a few tens of tokens.

\begin{figure}
    \centering
    \includegraphics[scale=0.22]{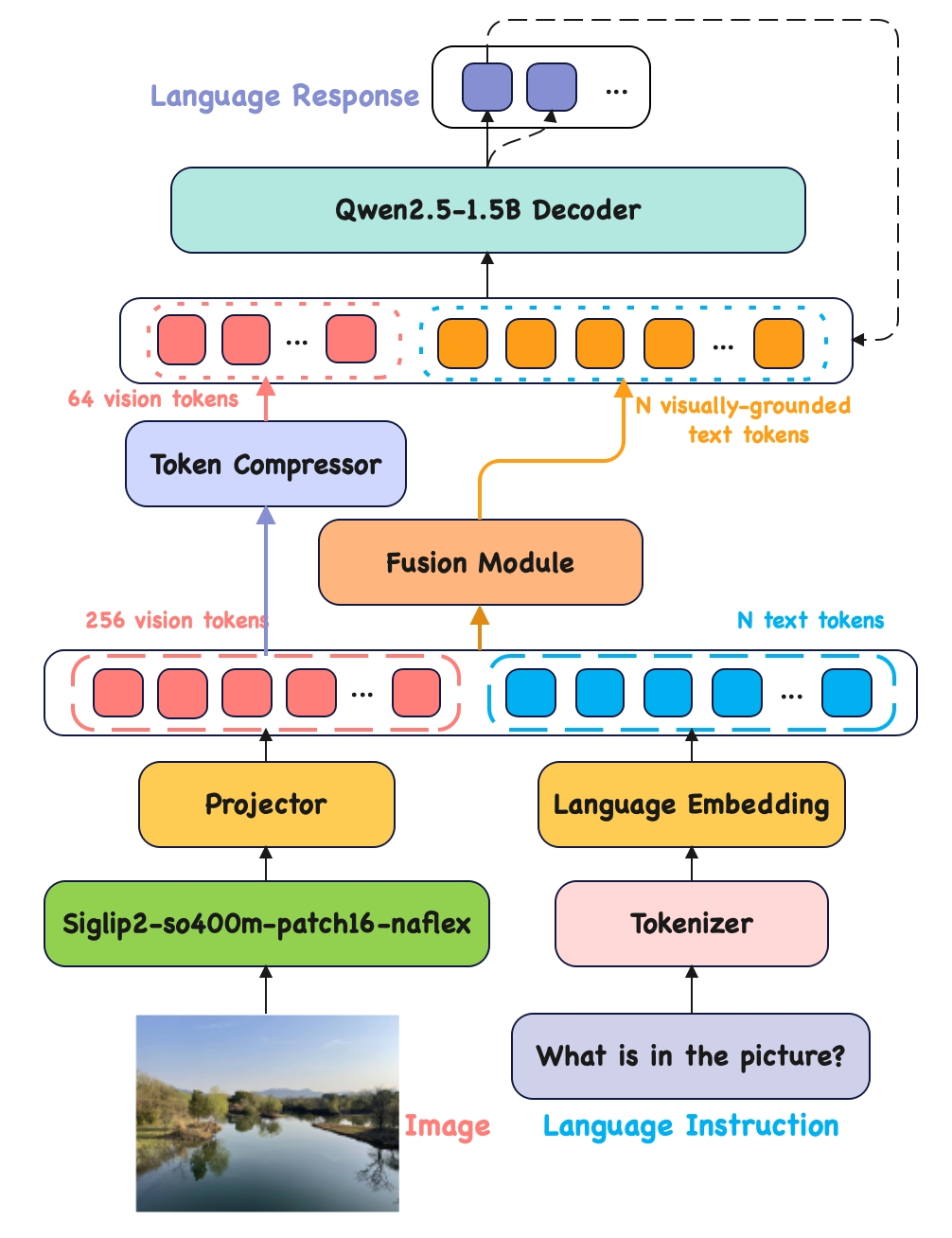}
    \caption{The overall architecture of AuroraEdge-V-2B. Based on LLaVA, we added a \textbf{Token Compressor} to reduce the number of visual tokens, and incorporated a \textbf{Fusion Module} to compensate for the visual loss caused by compression, thereby enhancing the efficiency of the model.The details of the  \textbf{Token Compressor}(see Figure \ref{fig:compressor}) and  \textbf{Fusion Module}(see Figure \ref{fig:fusion}) are introduced in the chapters below.}
    \label{fig:arch_01}
\end{figure}

In this paper, we introduce AuroraEdge-V-2B, a compact, stronger, faster, and more suitable VLLM for edge depolyment. 
We also propose a \textbf{compression-fusion} method to improve its inference efficiency. 
This method compresses the visual tokens while employing fusion techniques for visual compensation, significantly 
reducing the number of floating-point operations during inference while still ensuring its robust performance.
In Summary, our contributions to the community are as follows:
\begin{enumerate}[leftmargin=*, labelsep=0.5em, labelwidth=1.5em]
\item To deploy VLLMs in industrial applications, we introduce AuroraEdge-V-2B, an efficient VLLM with strong performance. 
In comparison with models of similar scale, it exhibits faster inference speed and achieves state-of-the-art (SOTA) performance on \textbf{9} benchmarks.
\item We propose a method for rapidly constructing VLLMs using pretrained LLMs and pretrained VEs. The method is highly 
reproducible and data-efficient, enabling us to rapidly build compact VLLMs we require.
\item To make VLLMs run faster, we propose a \textbf{compression-fusion} method to enhance inference efficiency. 
This method uses a token compressor to reduce the number of visual tokens, and simultaneously uses a fusion module 
to infuse visual information into text tokens, creating \textbf{visually-grounded text tokens} that substitute for the original ones, 
thereby compensating for the visual information loss from compression. As a result, our model not only achieves superior performance but also exhibits high inference efficiency.
Compared to other state-of-the-art (SOTA) models with the same number 
of parameters, such as Qwen2-VL-2B\cite{ref18}, Qwen2.5-VL-3B\cite{ref19}, InternVL2.5-2B\cite{ref20}, our model is \textbf{3} times faster.
\end{enumerate}

%% file: sec/2_related.tex
\section{Related Works}
\subsection{VLLMs Based On LLaVA Architecture}
With the rapid development of LLMs\cite{ref2,ref6,ref7}, VLLMs\cite{ref1,ref3,ref4,ref5,ref47,ref48,ref49,ref50,ref51,ref52} constructed based on pre-trained LLMs have demonstrated increasingly strong capabilities in visual understanding and reasoning.
A effective method for constructing VLLMs based on LLMs is the LLaVA\cite{ref1,ref3,ref4,ref5} architecture(see Figure \ref{fig:llava_01}), which requires only the addition of three components to handle visual inputs to the LLM: \textbf{an image processor}, \textbf{a visual encoder}, and \textbf{a projector}.

\textbf{The image processor} is mainly used for preprocessing input images, including scaling the images to specific sizes and slicing 
images into multiple patches to support dynamic resolutions. \textbf{The vision encoder} is used to encode image inputs into visual tokens. 
These encoders are typically based on the vision transformer (VIT)\cite{ref21} architecture and are pre-trained on large datasets of image-text pairs. 
There is a lot of good work being done in the field of image pretraining, such as the CLIP\cite{ref8,ref9} which is a classic work on contrastive language–image pre-training, 
the SigLIP\cite{ref10} which uses Softmax to replace Sigmoid in the CLIP loss function, the NaViT\cite{ref22} which supports arbitrary resolutions through the Patch n’Pack method, the 
FlexiViT\cite{ref23} which supports arbitrary patch sizes, and the recently SigLIP2\cite{ref24} by Google, which simultaneously supports multiple 
languages, arbitrary resolutions, and arbitrary patch sizes. \textbf{The projector} 
is used to project visual tokens into language space, with common methods including linear projectors, Q-Former\cite{ref25}, and Perceiver Resampler\cite{ref26}. 
Many existing state-of-the-art VLLM models are built upon the LLaVA architecture, such as the Qwen2-VL series\cite{ref18}, Qwen2.5-VL series\cite{ref19}, and Intern2.5-VL series\cite{ref20}, etc.

However, the application of VLLMs in industrial 
environments still faces challenges. (1) VLLMs generally have a large number of parameters, requiring significant computational 
resources for deployment. (2) The pre-trained vision encoders extract a large number of visual tokens, resulting in high inference latency.

\subsection{Lightweight VLLMs}
As an increasing number of researchers are devoting their efforts to the miniaturization of LLMs\cite{ref2,ref6,ref7}, small LLMs with fewer parameters are gradually 
demonstrating increasingly robust capabilities. Notable examples include the Qwen2.5 series (1.5B, 3B, 7B)\cite{ref27}, the Gemma series (1B, 2B, 4B)\cite{ref28}, 
MiniCPM3-4B\cite{ref29}, and Phi-4-Mini (3.8B)\cite{ref30}. To facilitate the deployment of VLLMs on edge devices, researchers are exploring the construction of small VLLMs using small LLMs, 
with typical works include Qwen2-VL-2B\cite{ref18}, Qwen2.5-VL-3B\cite{ref19}, Qwen2.5-VL-7B\cite{ref19}, MiniCPM-V-2-6-8B\cite{ref26}, and InternVL-2.5-2B\cite{ref20}.
Small VLLMs are more suitable for deployment on edge devices and often run faster. 

However, unlike 
LLMs which only have text inputs, small VLLMs still introduce a large number of visual tokens for the decoder, which leads to more floating-point 
operations, making achieving real-time inference like traditional DLMs challenging.

\subsection{Inference Acceleration of VLLMs}
To enable real-time inference for VLLMs, researchers have explored various methods to enhance their inference efficiency. 
Currently, the mainstream approaches to improving VLLM inference efficiency include two primary strategies. The first 
involves further model compression, using techniques such as knowledge distillation to reduce the model’s size\cite{ref12,ref13,ref14,ref15}. The 
second strategy employs quantization\cite{ref16} technology, converting the model to 4-bit or 8-bit precision, and performing INT4 or 
INT8 inference to boost efficiency. 

Additionally, some researchers have attempted to reduce the number of input visual tokens. 
For instance, LLaVA-Mini\cite{ref17} analyzed that visual tokens in VLLMs play a significant role only in the first few decoder layers 
during the decoding process. Consequently, it compresses visual tokens into a single token for decoding process and 
has achieved competitive results compared to LLaVA-1.5\cite{ref4}.

%% file: sec/3_model.tex
\section{AuroraEdge-V-2B}
In this section, we introduce AuroraEdge-V-2B in detail. 

\subsection{Architecture}
The overall structure of the model is shown in Figure \ref{fig:arch_01}. AuroraEdge-V-2B is also constructed based on the LLaVA architecture, encompassing five key modules:  \textbf{a vision encoder}, \textbf{a projector}, \textbf{a token compressor}, \textbf{a fusion module}, and \textbf{a LLM backbone}.
Unlike the conventional LLaVA models (as shown in Figure \ref{fig:llava_01}), to effectively control the number of visual tokens, the dynamic resolution processor\cite{ref4} 
is removed, and a token compressor and a fusion module are added to the framework. To balance model performance and inference efficiency for edge 
deployment, we use the pretrained Qwen2.5-1.5B\cite{ref27} as the LLM backbone.

Assuming \(X_v\) represent the visual inputs processed by the image processor, respectively, denotes \(X_t\) represent the text inputs processed by the tokenizer, 
\(F_{ve}\) denotes the visual encoder, \(F_{proj}\) denotes the projector, \(F_{embeds}\) denotes the text embedding layer, 
\(F_{compress}\) denotes the token compressor, \(H_{v}\) represents the visual tokens, \(H_{v2t}\) represents the visual tokens projected to language space, 
\(H_{vc}\) represents the compressed visual tokens, \(H_t\) represents the text tokens, \(H_{ft}\) denotes the text tokens fused with visual tokens,
\(H_{m}\) denotes the fused text tokens merged with compressed visual tokens,
\(F_{fusion}\) denotes the fusion module, \(F_{decode}\) denotes the LLM decoder, \(F_{cat}\) denotes feature concatenation, 
and \(Y\) represents the outputs, the inference process of AuroraEdge-V-2B is as follows: \\
\textbf{(1) Generate text tokens:} 
\begin{flalign} &\ \qquad H_{t}=F_{embeds}(X_t), \quad H_t \in R^{N_t \times D_t} & \end{flalign}
\textbf{(2) Generate visual tokens:} 
\begin{flalign}  &\ \qquad H_{v}=F_{ve}(X_v),\quad H_v \in R^{N_v \times D_v} & \end{flalign}
\textbf{(3) Project visual tokens into language space:} 
\begin{flalign} &\ \qquad H_{v2t}=F_{proj}(H_v), \quad H_{v2t} \in R^{N_v \times D_t}&
\end{flalign}
\textbf{(4) Compress visual tokens:} 
\begin{flalign} &\ \qquad H_{vc}=F_{compress}(H_{v2t}), \quad H_{vc} \in R^{N_c \times D_t}& 
\end{flalign}
\textbf{(5) Fuse text and visual tokens:} 
\begin{flalign} &\ \qquad H_{ft}=F_{fusion}(H_{v2t}, H_t), \quad H_{ft} \in R^{N_t \times D_t}& 
\end{flalign}
\textbf{(6) Merge text and compressed visual tokens:} 
\begin{flalign} &\ \qquad H_m=F_{cat}((H_{vc}, H_{ft}),\dim=0)& 
\end{flalign} 
\textbf{(7) LLM Decode:} 
\begin{flalign} &\ \qquad Y=F_{decode}(H_m)& 
\end{flalign}

\subsection{The Vision Encoder}
We employ \textbf{Siglip2-so400m-patch16-naflex}\cite{ref24} as the pre-trained visual encoder to support dynamic resolution in AuroraEdge-V-2B. 
This VE employs the Patch n’Pack method(NaViT \cite{ref22}) and randomizes the patch size(FlexiViT\cite{ref23}) during training, thus supporting dynamic patch sizes and sequence lengths.
It can efficiently control the number of visual tokens by predefining maximum number of patches for each image. The maximum number of patches for each image is set to 256, 
with a patch size of 16, meaning that the number of tokens extracted from the image by the VE is at most 256. When the number of 
extracted tokens is less than 256, the output sequence would be padded with zeros to reach a length of 256, and the positions of the padding would 
be marked as zero in the attention mask.
\subsection{The Projector}
Unlike the complex structure of projectors in other models, we employs a 2-layer MLP as the projector to map visual tokens to text space, 
which has been proven to be surprisingly powerful and data-efficient in LLaVA-1.5\cite{ref4}.
\subsection{The Token Compressor}
\begin{figure}
    \raggedright
    \includegraphics[scale=0.22]{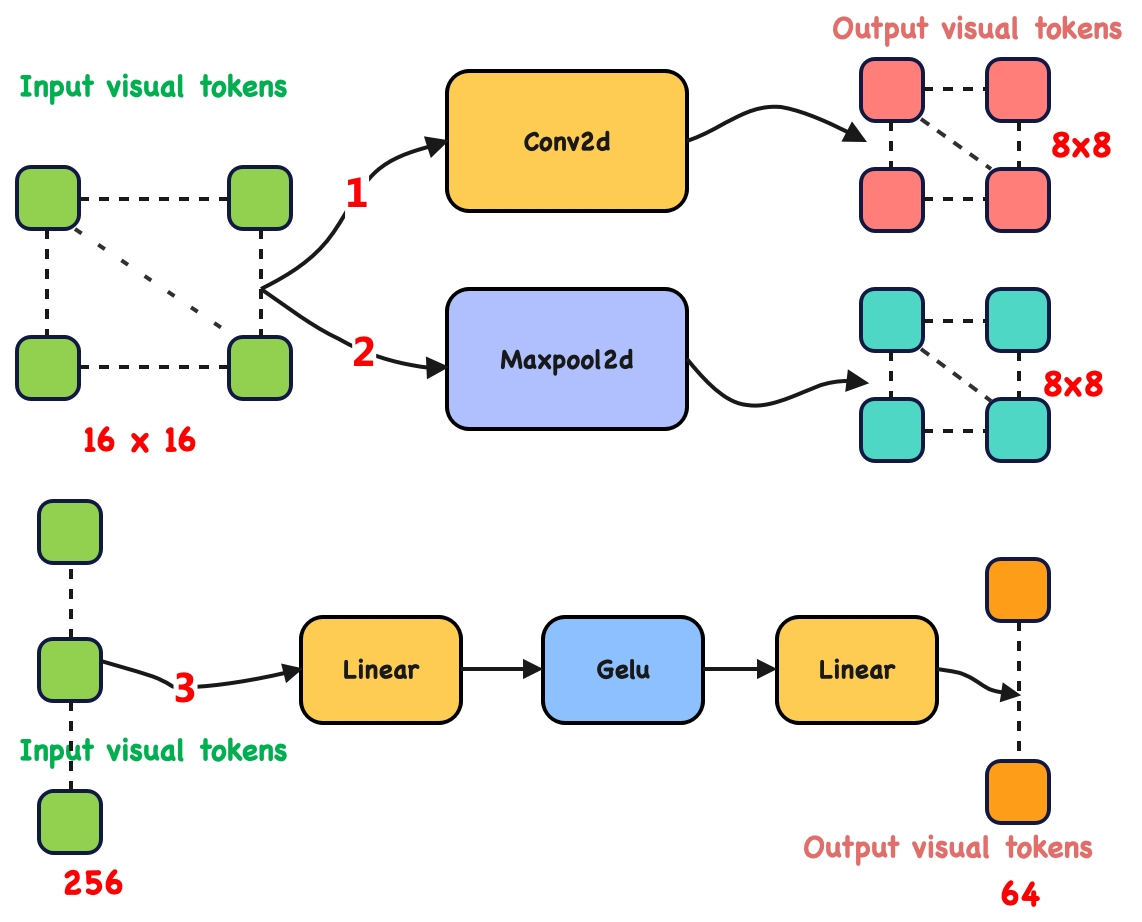}
    \caption{The token compressors of AuroraEdge-V-2B. We experimented with three methods—1.Conv2d, 2.MaxPool2d, and 3.MLP—for compressing the visual tokens. 
    Through training experiments on these methods, we observed that the MLP converges the fastest and achieves the best performance for the same number of training steps, 
    leading to its adoption as our final token compressor.}
    \label{fig:compressor}
\end{figure}
To further enhance the model’s inference efficiency, we incorporates a token compressor module designed to reduce the number 
of visual tokens. After processing through the visual encoder and projector, 
each image yields 256 visual tokens \(H_{v2t} \in R^{256 \times D_t}\), which are then mapped to a fixed number of visual tokens (we use 64) by the compressor. During the 
training process, we employs three methods for token compression: Convolution (Conv2d), MaxPooling (MaxPool2d), and 
Multi-Layer Perceptron (MLP). For details, please refer to Figure \ref{fig:compressor}. Experimental results demonstrate that compression via MLP yields superior model performance.
\subsection{The Fusion Module}
\begin{figure}
    \centering
    \includegraphics[scale=0.25]{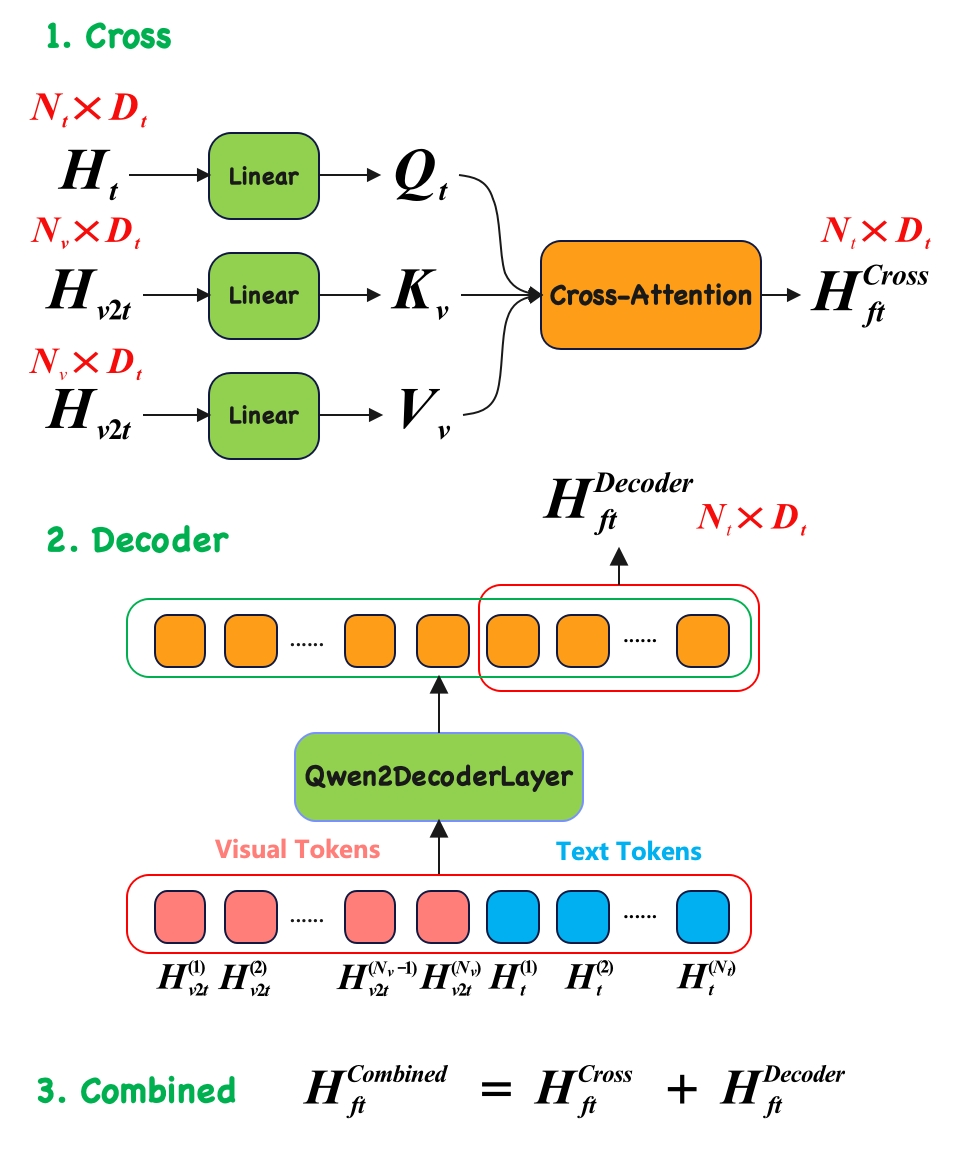}
    \caption{The fusion modules of AuroraEdge-V-2B. We experimented with three methods for multimodal fusion: 1. \textbf{Cross (Cross-Attention)}, 2. \textbf{Decoder} (A single-layer transformer decoder), 3. \textbf{Combined} (Accumulation of the results from Cross and Decoder). Among these approaches, 
    the model performance achieved by method 3 (\textbf{Combined}) was the best, and we adopt it as our final solution.}
    \label{fig:fusion}
\end{figure}
To compensate for the loss of visual information due to token compression, we have designed a fusion module that 
injects the original visual information into the text tokens. During the training process, we employs three methods 
for multimodal feature fusion: Cross-attention\cite{ref31} (\textbf{Cross}), a single-layer transformer decoder (\textbf{Decoder}), and the combination of the \textbf{Cross} and the \textbf{Decoder} (\textbf{Combined}).
(Refer to Figure \ref{fig:fusion}.) Experimental results in Table \ref{tab:evaluation_details} indicate that the \textbf{Combined} yields superior performance. 
The details of the three methods are as follows: \(H_t\) denotes the text tokens, 
\(H_{v2t}\) denotes the visual tokens before compression, \(F_{\text{Cross}}\) denotes the Cross method, \(F_{\text{Decoder}}\) denotes the Decoder method.\\
\\\textbf{(1) Cross:}
\begin{flalign} &\ Q_t = W^Q \cdot H_t \\
&\ K_v=W^K\cdot H_{v2t}, \quad  V_v=W^V\cdot H_{v2t} \\
&\ H_{ft}=\text{Softmax}(\frac{Q_t \cdot K_v^T}{\sqrt{D_t}}, \dim=1)\cdot V_v& \end{flalign} 
\textbf{(2) Decoder:}
\begin{flalign} & H_{cat}= \text{Concat}((H_{v2t}, H_t), \dim=0)  \\
& Q_{mix} = W^Q \cdot H_{cat}, \quad K_{mix}=W^K\cdot H_{cat}  \\
& V_{mix}=W^V\cdot H_{cat} \\
& H_{mix} =\text{Softmax}(\frac{Q_{mix} \cdot K_{mix}^T}{\sqrt{D_t}}, \dim=1)\cdot V_{mix} \\
& H_{ft}=H_{mix}\bigl[ N_v : \, \bigr]& \end{flalign} 
\textbf{(3) Combined:}
\begin{flalign} &\ H_{ft} = F_{\text{Cross}}(H_{v2t}, H_t) + F_{\text{Decoder}}(H_{v2t}, H_t)& \end{flalign}

%% file: sec/4_imple.tex
\section{Implementation details}
In this section, we will provide a detailed introduction to the implementation details of AuroraEdge-V-2B. To ensure the model achieves superior 
performance, we have collected a substantial amount of open-source datasets and simultaneously synthesized a significant quantity of training data.

\subsection{Initialization}
For initialization, we employ the pretrained Qwen2.5-1.5B\cite{ref27} for our LLM and the pretrained Siglip2-so400m-patch16-naflex\cite{ref24} for our Visual Encoder (VE).
The weights in the projector, the compressor and the cross-attention layer within the fusion module 
were initialized with random values from a standard normal distribution, while the weights in the transformer decoder within the fusion module were initialized
with the weights of the first decoder layer in the LLM backbone.

\subsection{Training Process}
After the initialization mentioned above, the entire training process is divided into three stages: \textbf{Vision Training}, \textbf{LLM Finetuning}, and \textbf{Vision-text Joint Training}. 
\\\textbf{Stage 1:Vision Training.} This phase primarily utilizes a large amount of image-caption data to train the modal connectors, including 
the projector, the token compressor, and the fusion module. During this stage, the parameters of the VE and the LLM are frozen, and 
only the parameters of the connectors are updated. Upon completion of this training phase, the connectors are capable of effectively aligning the image and text modalities, 
resulting in the model exhibiting strong performance on image captioning tasks.
\\\textbf{Stage 2:LLM Finetuning.} This phase mainly uses a large amount of visual question answering (VQA) data to enhance the model’s comprehension 
of the visual inputs. During the training process, the parameters of the VE are frozen, and all other modules except the VE are updated, 
including the projector, the compressor, the fusion module, and the LLM backbone. Upon completion of this training stage, the LLM’s understanding of mixed image-text tokens is significantly improved, 
resulting in robust performance on VQA tasks.
\\\textbf{Stage 3:Vision-text Joint Training.} This phase is primarily dedicated for joint image-text training, utilizing detailed image-caption data 
and complex VQA data to further improve the model's performance. During the phase, All parameters of the model are updated. 

\subsection{Data}
We employed a large number of image-caption data and a variety of VQA (Visual Question Answering) data to train our model. The open-source datasets used 
include LLaVA-Pretrain\cite{ref4}, MMDU\cite{ref33}, Flickr30k\cite{ref34}, COCO\cite{ref35} for image captioning, and AI2Diagram\cite{ref36}, OCRVQA\cite{ref43}, 
GQA\cite{ref37}, OKVQA\cite{ref45}, VQAV2\cite{ref39}, LLaVA-Instruct-150k\cite{ref4}, ShareGPT4V\cite{ref46} for VQA. Due to the scarcity of open-source image-text data, 
these datasets are also commonly shared in the training processes of most existing state-of-the-art (SOTA) models.

In addition, to augment our dataset with high-quality samples, we employed a synthesis pipeline. This pipeline leveraged images from the aforementioned open-source 
datasets, utilized a GPT model for text generation, and incorporated a manual curation step to ensure quality.

Through multiple experimental validations, we have summarized the following data preparation methods that are beneficial for training: 
\begin{enumerate}[leftmargin=*, labelsep=0.5em, labelwidth=1.5em]
    \item Image-caption data lacks explicit queries, so when constructing samples, it is advisable to use various questioning methods to generate diverse samples. For example, For instance, “Please describe the image” can be rephrased as “What is in the picture?” and similar operations to expand sample diversity.
    \item The answers in VQA data are usually brief. It is preferable to consolidate multiple QA data for the same image into a single sample.
    \item To enhance the model’s ability to generate longer texts, it is necessary to mix in a small amount of detailed image-caption data during Stage 2.
\end{enumerate}

\subsection{Hyperparameters}
We trained our model on a server equipped with eight NVIDIA A40 GPUs, each with 48 GB of memory.
The hyperparameters of the three training stages are shown in Table \ref{tab:training_details} 
and the three-stage training strategy are shown in Table \ref{tab:components_details}.

\begin{table}[h]
    \small
    \centering
    \renewcommand{\arraystretch}{1.2}
    \caption{The hyperparameters of AuroraEdge-V-2B during training.}
    \label{tab:training_details}
    \begin{tabular}{lccc}
        \hline
        & \textbf{Stage 1} & \textbf{Stage 2} &\textbf{Stage 3} \\
        \hline
        Batch Size & 32 & 32 & 16\\
        Epochs & 3 & 3 & 2 \\
        Learning Rate & 1e-3 & 2e-5 & 1e-6\\
        Schedule & Cosine decay & Cosine decay & Cosine decay \\
        Warmup Ratio & 0.03 & 0.03 & 0.03 \\
        Optimizer & AdamW & AdamW & AdamW \\
        \hline
    \end{tabular}
\end{table}

\begin{table}[h]
    \small
    \centering
    \renewcommand{\arraystretch}{1.2}
    \caption{The three-stage training strategy of AuroraEdge-V-2B. In Stage 1, we freeze the LLM and VE, training only the projector, 
    compressor, and fusion module. In Stage 2, we freeze the VE but train the LLM and the other modules. In Stage 3, we perform 
    full fine-tuning on all parameters with a low learning rate.}
    \label{tab:components_details}
    \begin{tabular}{lccc}
        \hline
        & \textbf{Stage 1} & \textbf{Stage 2} &\textbf{Stage 3} \\
        \hline
        Visual Encoder & Frozen & Frozen & Trainable \\
        Projector & Trainable & Trainable & Trainable \\
        LLM & Frozen & Trainable & Trainable \\
        Token Compressor & Trainable & Trainable & Trainable \\
        Fusion Module & Trainable & Trainable & Trainable \\
        \hline
    \end{tabular}
\end{table}

%% file: sec/5_result.tex
\section{Empirical Results}
To rigorously assess the performance of our proposed model, we selected a collection of 11 widely-recognized benchmarks for evaluation. The selected benchmarks are:
ScienceQA\cite{ref44}, VQAV2\cite{ref39}, OKVQA\cite{ref45}, 
TextVQA\cite{ref40}, VIZWIZ\cite{ref42}, GQA\cite{ref37}, AI2Diagram\cite{ref36}, OCRVQA\cite{ref43}, MMBench(cn)\cite{ref38}, MMBench(en)\cite{ref38}, and MME\cite{ref41}. 

It is noteworthy that some of these datasets comprise both training and benchmarking components. Critically, the image distributions of their training and test partitions are entirely disjoint. 
This structural separation ensures that training on the former does not invalidate the fairness of the evaluation on the latter.

We also evaluated other well-recognized 
models with similar parameter counts using the same evaluation scripts, including Qwen2.5-VL-3B\cite{ref19}, Qwen2-VL-2B\cite{ref18}, and InternVL-2.5-2B\cite{ref20}. 
Through comparison, we found that AuroraEdge-V-2B achieves higher scores on 9 of the 11 benchmarks. For specific details, please refer to Table\ref{tab:score_details}.

\begin{table}[h]
    \small
    \centering
    \renewcommand{\arraystretch}{1.2}
    \caption{The scores of AuroraEdge-V-2B on 11 benchmarks.In this table, Qwen2.5 refers to Qwen2.5-VL-3B, Qwen2 represents Qwen2-VL-2B, Intern2.5 stands for InternVL-2.5-2B, and AuroraEdge indicates AuroraEdge-V-2B. Out model (AuroraEdge) achieved state-of-the-art results across nine benchmarks(VQAV2, OKVQA, TextVQA, VIZWIZ, GQA, AI2Diagram, MME, MMBench(cn), MMBench(en)). }
    \label{tab:score_details}
    \begin{tabular}{l|cccc}
        \hline
        \multirow{2}{*}{\makecell{\textbf{Benchmarks}}} & \multicolumn{4}{c}{\textbf{Models}}  \\
        \cline{2-5}
        &\makebox[0.06\textwidth][c]{Qwen2.5} & \makebox[0.05\textwidth][c]{Qwen2} & \makebox[0.06\textwidth][c]{Intern2.5} & \makebox[0.07\textwidth][c]{AuroraEdge} \\
        \hline
        ScienceQA&80.49 & 77.59 & 95.80 &76.74\\
        VQAV2&80.51 & 80.01 & 75.28 & \textbf{83.21}\\
        OKVQA&55.22 & 52.96 & 46.58 & \textbf{73.95} \\
        TextVQA&76.82 & 76.87 & 69.44 & \textbf{77.26} \\
        VIZWIZ&70.35 & 66.96 & 45.76 & \textbf{71.05} \\
        GQA&61.25 & 60.4 & 59.3 & \textbf{65.75} \\
        AI2Diagram&60.85 & 55.35 & 73.55 & \textbf{92} \\
        OCRVQA&74.53 & 74.17 & 32.02 & 68.15 \\
        MMBench(cn)&82.98 & 74.24 & 74.29 & \textbf{92.39} \\
        MMBench(en)&82.07 & 73.36 & 71.76 & \textbf{83.72} \\
        MME&85.72 & 87.99 & 80.79 & \textbf{94.06} \\
        \hline
    \end{tabular}
\end{table}

Furthermore, to validate the model’s efficiency in real-world scenarios, we selected  
a single NVIDIA GeForce RTX 3090 GPU with 24GB of memory, a card commonly used in industrial applications, as our test machine, 
and used an image with a resolution of $640\times480$ as the visual input, along with the text input ``please describe the image.'' to test the model’s
floating-point operations and inference latency. We found our model exhibits fewer floating-point operations (less than 16\%) and demonstrates 
an advantage of more than 3 times in inference efficiency. For further details, refer to Table \ref{tab:evaluation_details}.

\begin{table}[h]
    \small
    \centering
    \renewcommand{\arraystretch}{1.2}
    \caption{The inference latency of AuroraEdge-V-2B. Throughout the tables below, “Qwen2.5” denotes Qwen2.5-VL-3B, “Qwen2” denotes Qwen2-VL-2B, “Intern2.5” denotes Intern2.5-VL-2B, 
    and “AuroraEdge” stands for AuroraEdge-V-2B.
    Our model is not only nearly three times faster but also requires less than 16\% of the FLOPs compared to the most computationally efficient baseline.}
    \label{tab:evaluation_details}
    \begin{tabular}{l|cccc}
        \hline
        \multirow{2}{*}{\textbf{Indicators}} & \multicolumn{4}{c}{\textbf{Models}}  \\
        \cline{2-5}
        &\makebox[0.05\textwidth][c]{Qwen2.5} & \makebox[0.05\textwidth][c]{Qwen2} & \makebox[0.05\textwidth][c]{Intern2.5} & \makebox[0.07\textwidth][c]{AuroraEdge} \\
        \hline
        Parameters&3.50B & 2.06B & 1.88B & 1.90B \\
        GFLOPS&2288.7 & 1645.4 & 4091.9  & 263.8 \\
        Lantency(ms)&143 & 116 & 142 & 40\\
        \hline
    \end{tabular}
\end{table}

%% file: sec/6_ablation.tex
\section{Ablation Study}
To assess the contribution of our \textbf{compression-fusion} method, i.e., the \textbf{Combined} approach in Figure \ref{fig:fusion}, we trained three variants—the Compress, Cross, and Decoder models—for a comparative analysis.
All these comparative models were trained using the same training procedure, and the details of the \textbf{Compress}, \textbf{Cross}, and \textbf{Decoder} models are listed in Table \ref{tab:ablation_models_detail}.

\begin{table}[h]
    \small
    \centering
    \renewcommand{\arraystretch}{1.2}
    \caption{Architectural details of the four models (including our final model AuroraEdge.) designed for comparison. The Cross, Decoder, and Combined fusion methods are illustrated in Figure \ref{fig:fusion}. All the four models use the same VE, projector, LLM and 
    token compressor(if applicable). The \textbf{Compress} only compresses without fusion. The \textbf{Cross}, \textbf{Decoder}, and \textbf{AuroraEdge} all perform both compression and fusion, differing in their fusion methodologies.}
    \label{tab:ablation_models_detail}
    \begin{tabular}{l|cccc}
        \hline
        \multirow{2}{*}{\textbf{Components}} & \multicolumn{4}{c}{\textbf{Comparison Models}} \\
        \cline{2-5}
        & \makebox[0.05\textwidth][c]{Compress} & \makebox[0.005\textwidth][c]{Cross} & \makebox[0.005\textwidth][c]{Decoder} & \makebox[0.008\textwidth][c]{AuroraEdge}\\
        \hline
        Vision Encoder& \ding{51} & \ding{51} & \ding{51} & \ding{51}\\
        Projector& \ding{51} & \ding{51} & \ding{51} & \ding{51} \\
        Token Compressor& \ding{51} & \ding{51} & \ding{51} & \ding{51} \\
        Fusion Module& \ding{56} & Cross & Decoder & Combined\\
        LLM&\ding{51}&\ding{51}&\ding{51}&\ding{51}\\
        \hline
    \end{tabular}
\end{table}
The scores of these four models on the benchmarks are presented in Table \ref{tab:ablation_score}.
The experimental results validate our design choices. The \textbf{Compress}, serving as an ablation baseline without fusion, yields the lowest performance, 
confirming the necessity of the fusion stage. A comparison between the two fusion strategies reveals that the \textbf{Decoder} approach is marginally superior 
to the \textbf{Cross} approach. Most importantly, the \textbf{AuroraEdge}(-V-2B), our proposed combined approach, despite exhibiting variability on specific metrics, 
achieves the state-of-the-art overall performance.

\begin{table}[h]
    \small
    \centering
    \renewcommand{\arraystretch}{1.2}
    \caption{The performance of the Comparison Models across 11 Benchmarks. }
    \label{tab:ablation_score}
    \begin{tabular}{l|cccc}
        \hline
        \multirow{2}{*}{\makecell{\textbf{Benchmarks}}} & \multicolumn{4}{c}{\textbf{Comparison Models}}  \\
        \cline{2-5}
        &\makebox[0.06\textwidth][c]{Compress} & \makebox[0.05\textwidth][c]{Cross} & \makebox[0.06\textwidth][c]{Decoder} & \makebox[0.07\textwidth][c]{AuroraEdge} \\
        \hline
        ScienceQA& 68.59 & 73.27 & 75.29 &\textbf{76.74}\\
        VQAV2& 72.3 & 75.62 & 84.27 & \textbf{83.21}\\
        OKVQA& 58.67 & 70.23 & 72.58 & \textbf{73.95} \\
        TextVQA& 66.37 & 75.37 & 77.73 & \textbf{77.26} \\
        VIZWIZ& 60.52 & 70.31 & 65.32 & \textbf{71.05} \\
        GQA& 58.63 & 66.41 & 62.4 & \textbf{65.75} \\
        AI2Diagram& 82.25 & 88.57 & 92.23 & \textbf{92} \\
        OCRVQA&70.78 & 68.35 & 66.62 & \textbf{68.15} \\
        MMBench(cn)& 78.45 & 86.34 & 88.41 & \textbf{92.39} \\
        MMBench(en)& 71.23 & 82.2 & 81.73 & \textbf{83.72} \\
        MME&80.25 & 87.46 & 88.79 & \textbf{94.06} \\
        \hline
    \end{tabular}
\end{table}

The inference latency of the four models are shown in Table \ref{tab:comparison_latency}.
The efficiency results confirm our initial hypothesis. The \textbf{Compress} exhibits the minimal computational footprint and latency. In contrast, the \textbf{AuroraEdge} incurs the highest computational cost and latency. 
Crucially, the absolute increase in computational load is marginal, and the latency discrepancy is insubstantial.

\begin{table}[h]
    \small
    \centering
    \renewcommand{\arraystretch}{1.2}
    \caption{The Inference Latency and FLOPs of the comparison models.}
    \label{tab:comparison_latency}
    \begin{tabular}{l|cccc}
        \hline
        \multirow{2}{*}{\textbf{Indicators}} & \multicolumn{4}{c}{\textbf{Comparison Models}}  \\
        \cline{2-5}
        &\makebox[0.06\textwidth][c]{Compress} & \makebox[0.05\textwidth][c]{Cross} & \makebox[0.06\textwidth][c]{Decoder} & \makebox[0.07\textwidth][c]{AuroraEdge} \\
        \hline
        Parameters& 1.83B & 1.86B & 1.88B  & 1.90B \\
        GFLOPS& 248.6 & 250.5 & 261.9 & 263.8 \\
        Latency(ms)& 36 & 37 & 38 & 40\\
        \hline
    \end{tabular}
\end{table}

Therefore, considering both performance and inference latency, the \textbf{Combined} method is clearly the superior fusion approach. Consequently, it has been adopted as the architecture 
for our final model, \textbf{AuroraEdge-V-2B}.

%% file: sec/7_limit.tex
\section{Limitations and Future Work}
In this work, We focus on the development of a compact VLLM suitable for industrial deployment, and We 
also undertook a series of exploratory studies to navigate the trade-off between model performance and inference latency.

To save training costs, we preset some hyperparameters, such as fixing the number of compressed visual 
tokens at \textbf{64} and setting the number of decoder layers in the fusion module to \textbf{1}. 
While these approaches have been proven effective, there is 
considerable room for improvement. Future work will focus on the following aspects: 
\\\textbf{Higher Compression Ratio:} To improve the model’s inference efficiency, we can further reduce the number of visual tokens and explore more 
sophisticated methods for visual information compensation, such as using multi-layer transformer decoders for fusion. 
\\\textbf{More Effective Compression Methods:} Currently, the supervision for the model’s compression module during training only comes from 
textual information. Utilizing visual labels to supervise the training of the compression module is another promising approach. For instance, 
we can train a standalone visual token compressor by constructing a task that involves reconstructing the original image from 
compressed tokens. By supervising the compression process with original visual information, we aim to enhance the information density of the compressed tokens. 
\\\textbf{Video Support:} Since video samples were not included in the training process, AuroraEdge-V-2B currently does not support video inputs. To better serve industrial production environments, 
we plan to train our model on video inputs to enable video understanding.

%% file: sec/8_conclusion.tex
\section{Conclusion}
In this paper, we introduce AuroraEdge-V-2B and provide a detailed explanation of its implementation. To further enhance the model’s 
efficiency for industrial deployment, we also propose a fusion-compression method to reduce the model’s inference FLOPs without compromising performance. 
This method compresses visual tokens while employing multimodal fusion techniques to inject visual information before compression into text tokens for visual 
compensation. To ensure the model achieves good performance, we conducted a three-stage training process on a large amount of collected and synthesized 
training data: vision training, LLM fine-tuning, and vision-text joint training. 
Following training, our model surpasses other models of comparable parameter size, including Qwen2-VL-2B\cite{ref18}, InternVL-2.5-2B\cite{ref20}, and Qwen2.5-VL-3B\cite{ref19}, 
across 9 open-source benchmarks and demonstrates a threefold increase in inference speed.

Furthermore, through ablation studies, we have demonstrated the effectiveness of our proposed compression-fusion method. 
This establishes a foundation for our future work on further enhancing the model’s inference efficiency.
With the further development of LLMs, 
it is a clear trend that VLLMs will be used in industrial scenarios in the future. However, at present, there is still a significant disparity between 
VLLMs and traditional domain-specific DLMs in terms of usability (including accuracy and real-time performance). We hope our work will draw more 
attention to smaller VLLMs, thereby facilitating the practical application of VLLMs in industrial processes.

%% file: sec/X_suppl.tex
\clearpage
\setcounter{page}{1}
\maketitlesupplementary

\section*{Data Synthesize}

We leverage a selection of popular multimodal datasets to synthesize additional data. This is achieved by prompting GPT-4o 
with our custom-constructed instructions, thereby augmenting the original data pool.
The datasets we used are listed in Table \ref{tab:dataset_details}, and the \textbf{VQA} is short for Visual Question Answering.

\begin{table}[h]
    \footnotesize
    \centering
    \renewcommand{\arraystretch}{1.2}
    \caption{The open-source dataset for AuroraEdge-V-2B.}
    \label{tab:dataset_details}
    \begin{tabular}{c|c}
    \hline
    \textbf{Task type} & \textbf{Dataset Name} \\
    \hline
    \multirow{4}{*}{\textbf{Image Caption}} & 
    \textbf{LLaVA-Pretrain} \cite{ref4}\\
    &\textbf{MMDU} \cite{ref33}\\
    &\textbf{Flickr30k} \cite{ref34}\\
    &\textbf{COCO} \cite{ref35}\\
    \hline
    \multirow{7}{*}{\textbf{VQA}} & 
    \textbf{AI2Diagram} \cite{ref36}\\
    &\textbf{OCRVQA} \cite{ref43}\\
    &\textbf{GQA} \cite{ref37}\\
    &\textbf{OKVQA} \cite{ref45} \\
    &\textbf{VQAV2} \cite{ref39}\\
    &\textbf{LLaVA-Instruct-150k} \cite{ref4} \\
    &\textbf{ShareGPT4V} \cite{ref46}\\
    \hline
    \end{tabular}
\end{table}

To enrich the training data, we design four 
distinct strategies to synthesize additional training samples.
Our methodology can be categorized along two dimensions: the output format (Caption vs. VQA) and the verbosity level (Brief vs. Detailed). The four resulting strategies are:

\begin{enumerate}[leftmargin=*, labelsep=0.5em, labelwidth=1.5em, itemsep=5pt]
    \item \textbf{Brief-Caption:} Synthesizing concise image descriptions. 
    We construct a brief-caption dataset by leveraging images from the MMDU\cite{ref33} and Flickr30k\cite{ref34} dataset. First, we formulate a diverse set of short descriptive instructions A few examples are shown in Table \ref{tab:brief_caption}. 
    For each image, an instruction is randomly selected from the set of short descriptive instructions and fed into GPT-4o to generate a corresponding image-text sample. 
    This process yields the final Brief-Caption dataset, which is then merged with the original caption dataset. To ensure consistency, for the original image-text 
    pairs that lack instructions, we also randomly select a descriptive instruction for each.

    \item \textbf{Detailed-Caption:} Synthesizing comprehensive image descriptions. Similarly, to the brief-caption data generation, we construct a set of detailed description instructions, 
    Table \ref{tab:detailed_caption} provides a few examples.
    These instructions are then used with a subset of images from the AI2Diagram\cite{ref36}, GQA\cite{ref37}, and LLaVA-Instruct-150k\cite{ref4} to generate a detailed description dataset.
    
    \item \textbf{Brief-VQA:} Generating concise question-answer pairs about the image. To construct the brief VQA dataset, we leverage the instructions outlined in Table \ref{tab:brief_vqa} .
     and apply them to a subset of images from the LLaVA-Pretrain\cite{ref4} and OCRVQA\cite{ref43}. The resulting dataset is subsequently integrated with other VQA datasets.

    \item \textbf{Detailed-VQA:} Generating comprehensive question-answer pairs about the image. To enhance the model’s comprehensive perception of images, we also generate a detailed VQA dataset. 
     This is achieved by using a selection of images from the AI2Diagram\cite{ref36}, GQA\cite{ref37}, and COCO\cite{ref35} datasets, along with the instructions provided in Table \ref{tab:detailed_vqa}.
\end{enumerate}

\begin{table}[h]
    \small
    \centering
    \renewcommand{\arraystretch}{1.2}
    \caption{Instructions for brief image caption. These instructions are designed to elicit holistic descriptions of images. 
    The central objective is to prompt GPT-4o to generate concise summaries of the visual content.}
    \label{tab:brief_caption}
    \begin{tabular}{|l|}
    \hline
    \textbf{Brief Caption Instructions} \\
    \hline
    Describe the image concisely.  \\ \hline
    Give a brief description of the image.  \\ \hline
    Give a short and clear explanation of the subsequent image. \\ \hline
    Present a compact description of the photo's key features.  \\ \hline
    Provide a brief description of the given image. \\ \hline
    Share a concise interpretation of the image provided.  \\ \hline
    Summarize the visual content of the image.  \\ \hline
    Write a terse but informative summary of the picture.   \\ \hline
    ...... \\ 
    \hline   
    \end{tabular}
\end{table}

\begin{table}[h]
    \small
    \centering
    \renewcommand{\arraystretch}{1.2}
    \caption{Instructions for detailed image caption. These instructions primarily focus on a fine-grained analysis of the image content to generate more detailed image understanding data.}
    \label{tab:detailed_caption}
    \begin{tabular}{|l|}
    \hline
    \textbf{Detailed Caption Instructions} \\
    \hline
    Describe the following image in detail. \\ \hline
    Provide a detailed description of the given image. \\ \hline
    Give an elaborate explanation of the image you see.\\ \hline
    Offer a thorough analysis of the image. \\ \hline
    Explain the various aspects of the image before you. \\ \hline
    Clarify the contents of the displayed image with great detail. \\ \hline
    Characterize the image using a well-detailed description. \\ \hline
    Write an exhaustive depiction of the given image. \\ \hline
    ...... \\ \hline   
    \end{tabular}
\end{table}

\begin{table}[h]
    \small
    \centering
    \renewcommand{\arraystretch}{1.2}
    \caption{Instructions for brief VQA. This instruction is primarily used to randomly extract two sets of question-answer (QA) data 
    from an image and output them in a structured format, thereby enriching the QA dataset.}
    \label{tab:brief_vqa}
    \begin{tabular}{|p{\linewidth}|}
    \hline
    \textbf{Brief VQA Instructions} \\
    \hline
    Please help me extract two pairs of question and answer from the image, the outputs must be in JSON format, using “Query” and “Answer” as fields.
    \\ \hline   
    \end{tabular}
\end{table}

\begin{table}[h]
    \small
    \centering
    \renewcommand{\arraystretch}{1.2}
    \caption{Instructions for brief VQA. This instruction is designed to guide GPT-4o in performing a multi-faceted analysis of the input image. 
    This process yields more comprehensive and intricate QA data, ultimately enhancing the model’s comprehension of complex scenarios.}
    \label{tab:detailed_vqa}
    \begin{tabular}{|p{\linewidth}|}
    \hline
    \textbf{Detialed VQA Instructions} \\
    \hline
    You are an AI visual assistant, and you are seeing a single image. \\
    Design a conversation between you and a person asking about this photo. The answers should be in a tone that a visual AI assistant is seeing the image and answering the question.
    Ask diverse questions and give corresponding answers. \\
    Include questions asking about the visual content of the image, including the object types, counting the objects, object actions, object locations, 
    relative positions between objects, etc. Only include questions that have definite answers: \\
    (1) one can see the content in the image that the question asks about and can answer confidently; \\
    (2) one can determine confidently from the image that it is not in the image. \\
    Do not ask any question that cannot be answered confidently. \\
    Also include complex questions that are relevant to the content in the image, for example, asking about background knowledge of the objects in the image, 
    asking to discuss about events happening in the image, etc. Again, do not ask about uncertain details. \\
    Provide detailed answers when answering complex questions. For example, give detailed examples or reasoning steps to make the content more convincing and well-organized.  
    You can include multiple paragraphs if necessary. Finally, your need to answer the last question, which is “Provide a detailed description of the given image”. 
    The conversation outputs should be in JSON format, using “Query” and “Answer” as fields..
    \\ \hline   
    \end{tabular}
\end{table}

\subsection*{Training Data Details}
As described in the body of the paper, our training methodology comprises three phases: 
Vision Training, LLM Finetuning, and Vision-text Joint Training. The datasets utilized for each phase are detailed below:

\begin{enumerate}[leftmargin=*, labelsep=0.5em, labelwidth=1.5em, itemsep=5pt]
    \item \textbf{Vision Training:} 
    The first training stage leverages a combination of Flickr30K\cite{ref34}, LLaVA-Pretrain\cite{ref4}, COCO\cite{ref35}, MMDU\cite{ref33}, and our in-house synthesized Brief Caption dataset. 
    The primary objective of this phase is to train the multimodal alignment module with high-level, general-purpose image-text descriptions.
    \item \textbf{LLM Finetuning:} 
    The second training phase is designed to bolster the LLM’s ability to produce text grounded in visual information. To this end, 
    we employ a combination of open-source VQA datasets (e.g., AI2Diagram \cite{ref36}, OCRVQA\cite{ref43}, VQAV2\cite{ref38}, LLaVA-Instruct-150k\cite{ref42}, OKVQA\cite{ref39}, GQA\cite{ref40}, 
    and a subset of ShareGPT4V\cite{ref41}), alongside our in-house synthesized Brief-VQA and a part of Detailed-Caption dataset.
    \item \textbf{Vision-text Joint Training:} To further enhance the model’s fine-grained understanding of visual inputs while preserving its global overview capability, we conduct mixed 
    training using our synthesized Detailed-VQA data, a selection of Detailed-Caption data, and a selection of Brief-Caption data.
\end{enumerate}

\subsection*{Case Studies of Model Inference}

In this section, we will present the model’s inference performance across four tasks: brief caption, detailed caption, brief VQA, and detailed VQA.

\begin{figure}
    \centering
    \includegraphics[scale=0.4]{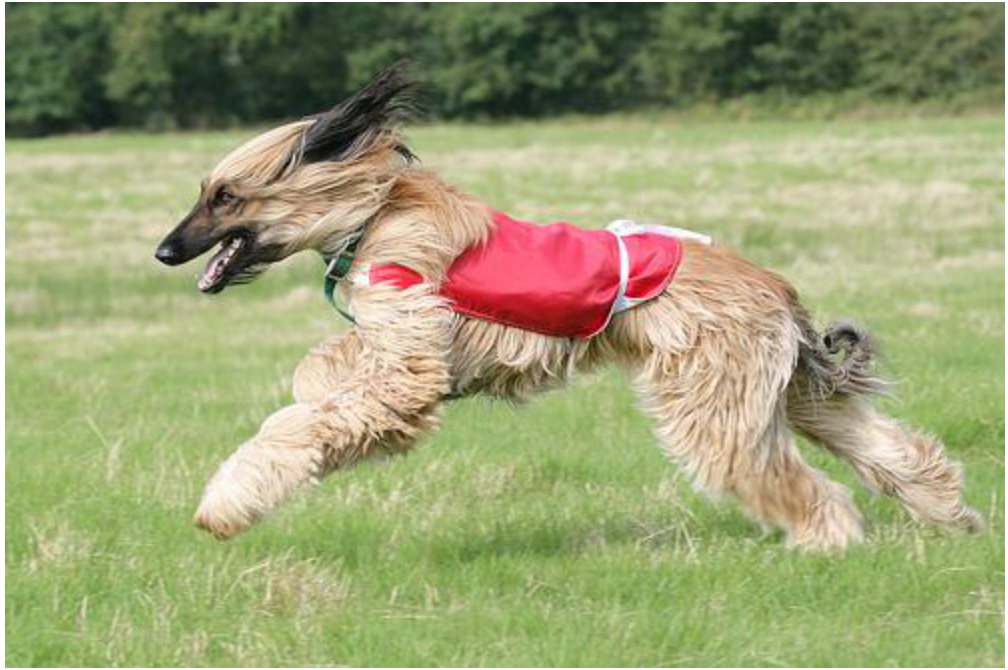}
    \caption{A lively dog is running across a lush green field.}
    \label{fig:case_1}
\end{figure}

\subsubsection*{Caption Case Study 1: }

Taking Figure \ref{fig:case_1} as the input, we executed the brief and detailed captioning tasks. 
The corresponding outputs of our model are presented below:

\begin{enumerate}[leftmargin=*, labelsep=0.5em, labelwidth=1.5em, itemsep=5pt]
    \item \textbf{Brief Caption Instruction:} “Give a brief description of the image."
    \item \textbf{Brief Caption Answer:} “The image depicts a dog with long, flowing hair running energetically across a grassy field."
    \item \textbf{Detailed Caption Instruction:} “Describe the following image in detail."
    \item \textbf{Detailed Caption Answer:} “In this dynamic image, a lively dog is captured mid-stride as it energetically runs across a lush green field. The dog has a long, silky 
coat of light tan fur that flows gracefully with its movement. It wears a bright red vest, which adds a vibrant splash of color against 
its fur. The dog's mouth is open, possibly panting or expressing excitement, and its ears are perked up, adding to the sense of motion 
and vitality. The background consists of a serene, open grassy area that gradually fades into a dense line of trees, suggesting a natural, 
outdoor setting. "
\end{enumerate}

\begin{figure}
    \centering
    \includegraphics[scale=0.05]{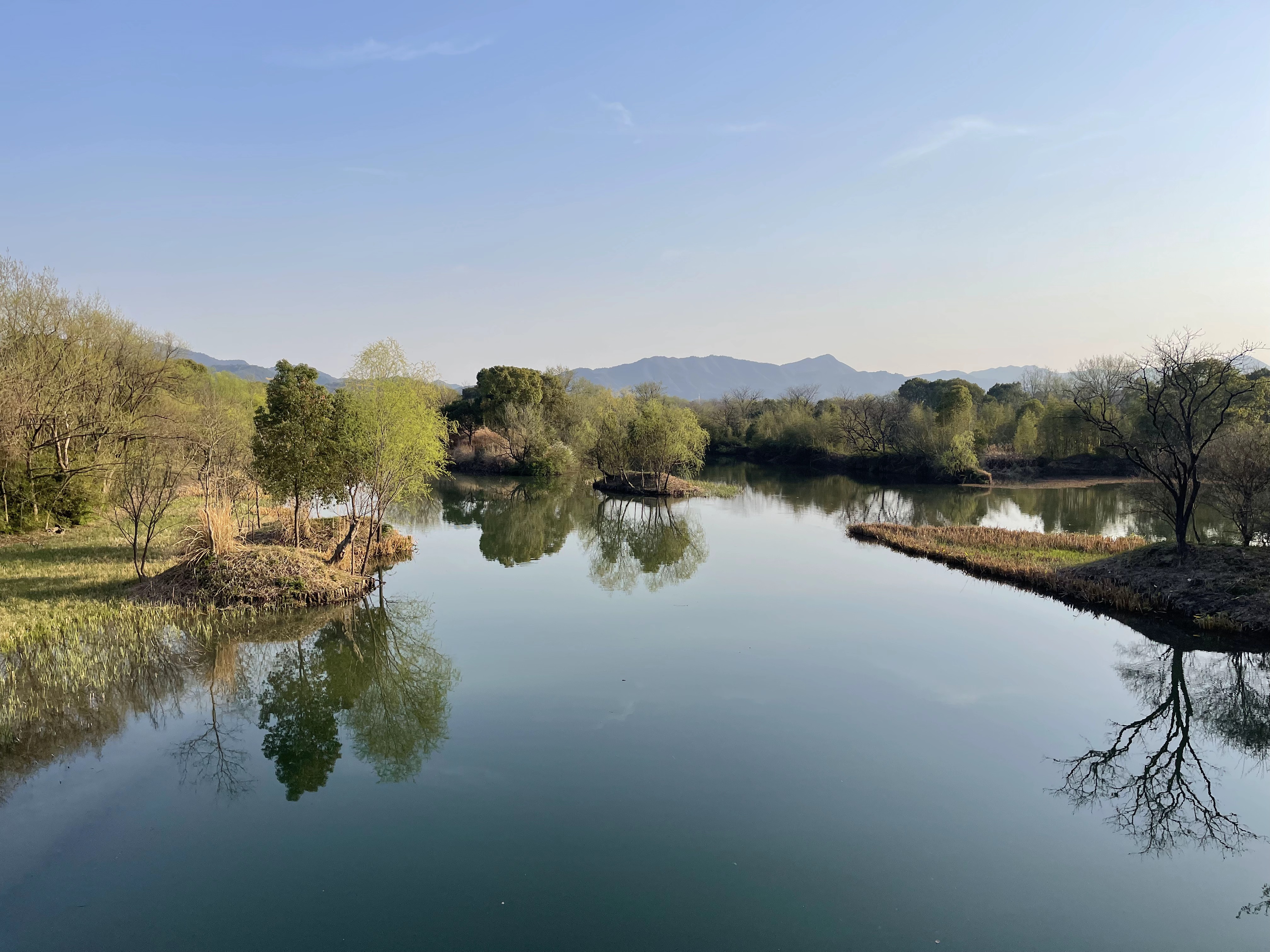}
    \caption{A pristine lake.}
    \label{fig:case_2}
\end{figure}

\subsubsection*{Caption Case Study 2: }
For the second caption case, we used Figure \ref{fig:case_2} as input. The inference result is as follows:
\begin{enumerate}[leftmargin=*, labelsep=0.5em, labelwidth=1.5em, itemsep=5pt]
    \item \textbf{Brief Caption Instruction:} “Present a compact description of the photo's key features."
    \item \textbf{Brief Caption Answer:} “A tranquil lake scene with calm, reflective water mirroring surrounding trees and distant mountains. 
    Small vegetated landmasses are scattered across the water under a clear blue sky."
    \item \textbf{Detailed Caption Instruction:} “Offer a thorough analysis of the image."
    \item \textbf{Detailed Caption Answer:} “The image depicts a tranquil natural landscape. The central feature is a body of calm water with a mirror-like surface, which clearly 
    reflects the surrounding trees and sky. The water is bordered by dense vegetation, with the foliage exhibiting a vibrant green, indicative of a lush growing season. 
    In the background, a continuous mountain range is visible, its contours sharply defined against the sky. The sky itself is a pale blue, almost devoid of clouds, creating 
    an impression of a clear and expansive atmosphere. The overall composition is characterized by rich colors and distinct layers, evoking a sense of tranquility and serenity. "
\end{enumerate}

\begin{figure}
    \centering
    \includegraphics[scale=0.6]{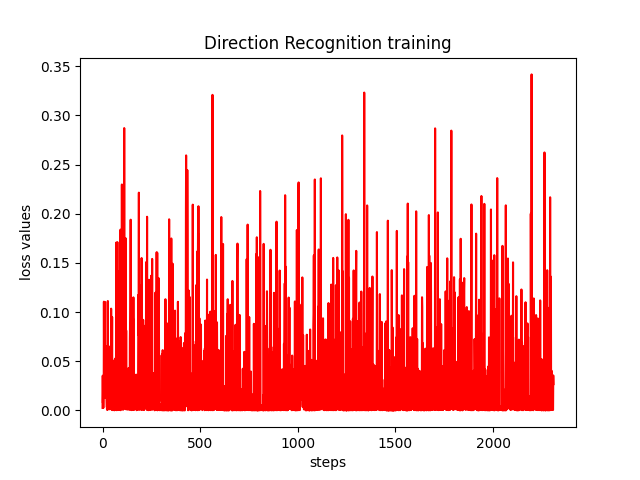}
    \caption{A Training Loss Curve.}
    \label{fig:case_3}    
\end{figure}

\subsubsection*{VQA Case Study 1: }
To evaluate the model’s performance on the Visual Question Answering (VQA) task, we begin with a case study using Figure \ref{fig:case_3}. The corresponding inference result is presented as follows:
\begin{enumerate}[leftmargin=*, labelsep=0.5em, labelwidth=1.5em, itemsep=5pt]
    \item \textbf{Brief VQA Instruction:} “What is the title of this figure?"
    \item \textbf{Brief VQA Answer:} “Direction Recognition Training." 
    \item \textbf{Detailed VQA Instruction:} “What do the x and y axes represent in this image, and what are their respective numerical ranges?"
    \item \textbf{Detailed VQA Answer:} “The x-axis represents "steps" with a numerical range from 0 to 2000.
The y-axis represents "loss values" with a numerical range from 0.00 to 0.35."
\end{enumerate}

\begin{figure}
    \centering
    \includegraphics[scale=0.6]{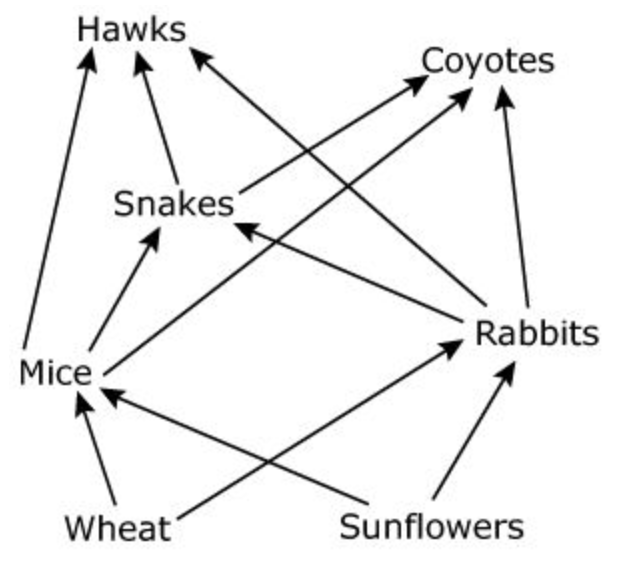}
    \caption{A Directed Connected Graph.}
    \label{fig:case_4}    
\end{figure}

\subsubsection*{VQA Case Study 2: }
To test the basic reasoning ability of our model, we selected a directed connected graph (see Figure \ref{fig:case_4}). The model’s inference result is shown below:
\begin{enumerate}[leftmargin=*, labelsep=0.5em, labelwidth=1.5em, itemsep=5pt]
    \item \textbf{Brief VQA Instruction:} “What animals eat mice?"
    \item \textbf{Brief VQA Answer:} “Hawks, Snakes, Coyotes."
    \item \textbf{Detailed VQA Instruction:} “How does energy flow from the producers to the top predators?"
    \item \textbf{Detailed VQA Answer:} “Energy flows from the producers (Wheat and Sunflowers) to the primary consumers
 (Mice and Rabbits) when they eat the plants. Then, the secondary consumers (Snakes) eat the 
primary consumers. Finally, the top predators (Hawks and Coyotes) eat both the primary consumers
 (Mice and Rabbits) and the secondary consumer (Snake). Each time an organism is eaten, energy 
is transferred up the food web."
\end{enumerate}

\subsection*{Concluding Remarks}

As can be clearly seen from the experimental results above, our model demonstrates strong capabilities on both the Image Captioning and VQA tasks. Furthermore, 
it exhibits a certain degree of reasoning ability in complex visual scenarios. Notably, 
we did not construct a large-scale dataset specifically for these reasoning scenarios, which represents a key direction for our future continuous optimization.